\documentclass{article}

 \usepackage[main, final]{neurips_2026}

\usepackage[utf8]{inputenc} 
\usepackage[T1]{fontenc}    
\usepackage{hyperref}       
\usepackage{url}            
\usepackage{booktabs}       
\usepackage{amsfonts}       
\usepackage{nicefrac}       
\usepackage{microtype}      
\usepackage{xcolor}         

\usepackage{amsmath}
\usepackage{graphicx}
\usepackage{multirow}

\usepackage{xcolor}
\definecolor{C0}{HTML}{1F77B4}
\definecolor{C1}{HTML}{FF7F0E}

\title{Learning a Maximum Entropy Model \\for Visual Textures using Diffusion}

\author{%
  Xinyuan Zhao \\
  New York University \\
  \texttt{xz2556@nyu.edu} \\
  \And
  Eero P. Simoncelli \\
  New York University \& Flatiron Institute \\
  \texttt{eero.simoncelli@nyu.edu} \\
}

\begin{document}

\maketitle

\begin{abstract}
Visual textures -- spatially homogeneous image regions  containing repeated elements (e.g. a field of grass, the bark of a tree) -- are ubiquitous in visual scenes and provide important cues for recognizing and analyzing materials, surface geometry and objects. A number of existing texture models extract essential statistics from a single texture image, and can then generate high-quality samples that are visually similar to the original by matching these statistics. However, their statistics are either hand-designed or based on a network pretrained for another purpose (e.g., object recognition). Here, we develop a principled method for unsupervised learning of a set of statistics that are used to constrain a maximum entropy probability model for texture. We leverage methods developed for generative diffusion models to derive training and sampling procedures, and compare these to the traditional method of sampling via matching the statistics. Despite the compactness of our trained model (512 statistics), it generates texture images whose quality is as good as or better than the current state-of-the-art model ($\sim$177k statistics), both visually and by quantitative perceptual scoring using DreamSim. Finally, we show that images generated from parameters lying along a line between the parameter vectors corresponding to two different reference textures are homogeneous textures with features that smoothly transition between those of the two references.
\end{abstract}

\section{Introduction}


Most photographic images contain ``visual textures'' - homogeneous regions of repeated elements, subject to some randomization in location, size, color, orientation, or other attributes. 
Visual texture representation and synthesis methods have been extensively studied \citep[e.g., ][]{heeger1995pyramid, de1997non, efros2001image, zhu1998filters,  portilla2000parametric,gatys2015texture,lu2016learning,ding2020image,de2021maximum}, and have served as a basis for style transfer \citep{gatys2016image}. A bias toward texture (as opposed to shape) contributes to the misalignment between machine and human object recognition \citep{geirhos2018imagenet}.  
Visual textures also provide potent stimuli for visual  neuroscience, where they have been used to differentiate the response selectivity of neurons in area V2 of the primate visual cortex relative to those in the preceding area V1 \citep{freeman2013functional,yu2015visual}. 

Here, we aim to develop a \emph{parametric} texture model, which takes a texture image as input and transforms it into a representation (a set of parameters or statistics), and in addition, a procedure for synthesizing new images of the same texture type based on this representation. Most existing parametric texture models are based on Julesz's seminal conjecture \citep{julesz1962visual}: the appearance of a texture is determined by a set of local statistics that are measured by the human visual system.  Two textures with the same statistics will have the same (or similar) appearance. This concept has been instantiated in a variety of texture models and their corresponding sampling procedures (e.g., \cite{heeger1995pyramid, zhu1998filters, portilla2000parametric,victor2012local,gatys2015texture}).
These models demonstrate their success by synthesizing images whose statistics match those estimated from an original texture image, and showing that these are similar in appearance to the original. However, the statistics they use are either hand-designed or based on a pretrained object-recognition network. 

The definition of a texture ensemble as a set of images with a given set of statistics can be formalized as a parametric probability model: images of a given texture class are samples drawn from the maximum entropy density \citep{jaynes1957information} (i.e., the ``most random'' density) subject to consistency with the values of the chosen statistics \citep{zhu1998filters, portilla2000parametric, victor2012local, lu2016learning, de2021maximum}. 
In these previous instantiations, the statistics were hand-selected.  Here, we develop a novel method for {\em unsupervised learning of a set of such statistics} from a dataset of texture images, and complementary methods for sampling from the maximum entropy density subject to these statistics.  In this context, the vector of these statistics computed on any texture image provides a parameterization of the conditional density of the corresponding texture class.

To learn a maximum entropy conditional density model, we make use of 
``generative diffusion'' methods that currently underlie most state-of-the-art image synthesis systems \citep{song2019generative, ho2020denoising,ramesh2022hierarchical,rombach2022high}. The main concept is simple: a network trained to remove additive Gaussian noise from images provides an estimate of the score (gradient of the log probability) of the noisy image distribution.  This gradient may then be used in an iterative reverse diffusion scheme to draw samples from a learned density of the training images.
Diffusion models have been used to solve linear inverse problems using conditional sampling
\citep{kadkhodaie2021stochastic, kawar2022denoising, chung2022diffusion, daras2024survey}.   The internal representation of denoisers used for diffusion has been shown to concentrate images with similar features in low-dimensional subspaces~\citep{Kadkhodaie26c}.
 More generally, they are often used to draw samples conditioned on auxiliary information such as class labels or text prompts. 
Diffusion models have also been used for the purpose of unsupervised representation learning (e.g., \cite{fuest2026diffusion}). 
In many cases, the conditioning signal is provided by a separate (nonlinear) image representation
\citep{mittal2023diffusion,preechakul2022diffusion,wang2023infodiffusion,hudson2024soda,yang2023lossy}. The conditioning and denoising networks are jointly trained on a set of images and their noisy counterparts. This configuration resembles an autoencoder, in which the representation network is the encoder, and the conditional denoiser is the decoder. 

Here, we develop a model of similar structure -- one network is used to compute a set of (learned) statistics, which are then used to condition a second network which performs denoising.
But unlike previous methods, the architectural and algorithmic implementation of the conditioning is dictated by the mathematical form of the maximum entropy model that we assume.
The primary contributions of the paper are:
\begin{itemize}
\item We develop a novel method for learning a set of statistics that can parameterize a maximum entropy density model for the training data.  To our knowledge, this is the first of its kind.
\item We train this model on a large ensemble of homogeneous image patches taken from ImageNet21K \citep{deng2009imagenet}, and show on a test set that the conditional samples from the model closely resemble the texture images on which they are conditioned. The results are visually similar or superior to the current state-of-the-art parametric texture model \citep{gatys2015texture}, while relying on a far smaller parameter set (512 vs.\ 176,640). We also achieve a better DreamSim score \citep{fu2023dreamsim} for a majority of the homogeneous textures used for testing.
\item We compare the conditional diffusion sampling procedure to the statistics matching procedure used to synthesize images in many previous texture models.  Although these are mathematically closely related, they yield different results, with statistics-matching samples generally of higher quality.
\item We examine the structure of the representation space by sampling from the model conditioned on parameter vectors interpolated between those of two different texture images.
We find that the model of \citet{gatys2015texture} generally produces images consisting of a patchwise spatial mixture of the two textures.  In contrast, our model generates homogeneous texture samples whose features lie between the two textures.
\end{itemize}

The code for our model (including trained weights) is publicly available\footnote{\url{https://github.com/Bichidian/maxent-texture}}.

\section{Method}

\subsection{Maximum entropy density model}
We aim to learn a parametric family of probability densities over $x\in\mathbb{R}^n$ (for an $n$ pixel image), where each density corresponds to a texture class, and individual texture images are samples. We assume a maximum entropy formulation \citep{jaynes1957information}, on which several existing texture models are based \citep{zhu1998filters, portilla2000parametric, lu2016learning, de2021maximum}. The maximum entropy probability distribution is the solution to the following optimization problem, parameterized by statistical values $\mu = [\mu_1,\mu_2,\cdots,\mu_d] \in \mathbb{R}^d$:
\begin{equation}\label{eq:maxent-problem}
    \max_p \mathbb{E}_p [-\log p(x)] \quad \text{s.t.} \ \ \mathbb{E}[f_k(x)] = \mu_k,\quad k=1,2,\cdots,d ,
\end{equation}
where $f: \mathbb{R}^n \to \mathbb{R}^d$ is a function that computes a vector of $d$ statistics. The optimal solution is an ``exponential family'' (see Appendix \ref{app:maxent} for the derivation):
\begin{equation}\label{eq:exponential-family}
    p_\lambda(x) = \frac{1}{Z(\lambda)} \exp \left( - \lambda^T f(x) \right) ,
\end{equation}
where $Z(\lambda)$ is a normalizing factor and $\lambda = [\lambda_1,\lambda_2,\cdots,\lambda_d] \in \mathbb{R}^d$ is chosen such that the density satisfies the constraints $ \mathbb{E}_{p_\lambda} [f_k(x)] = \mu_k $. The two $d$-dimensional vectors $\lambda$ and $\mu$ are uniquely determined by each other (Appendix \ref{app:maxent}), and either can be used as a parametric specification of the conditional density of a texture class. 

\subsection{Training a maximum entropy model via denoising}
Direct optimization of $f$ and $\lambda$ by maximizing likelihood over a training set is generally infeasible, since $Z(\lambda)$ requires integration over the entire data space. Instead, we propose to learn $f$ indirectly via denoising, using a formulation similar to that of generative diffusion models. For a texture image $x$, consider the noise-contaminated $y$, expressed using the variance-preserving formulation of  \cite{ho2020denoising}:
\begin{equation}
    y = \frac{1}{\sqrt{\sigma^2+1}} x + \frac{\sigma}{\sqrt{\sigma^2+1}} \varepsilon ,
\end{equation}
where $\varepsilon$ is a sample of Gaussian noise with zero mean and identity covariance. 
A function $\hat{\varepsilon}(y)$ trained to minimize the mean squared error in predicting $\varepsilon$ will compute an approximation of the posterior mean, $\mathbb{E}(\varepsilon|y)$. This conditional mean, in turn, is proportional to the score (gradient of the log probability) of the noisy density $p(y)$ \citep{robbins1956empirical,miyasawa1961empirical,raphan2011least} (proof in Appendix \ref{app:miyasawa}):
\begin{equation}\label{eq:miyasawa}
    \mathbb{E} (\varepsilon|y) = - \frac{\sigma}{\sqrt{\sigma^2 + 1}} \nabla_y \log p(y) .
\end{equation}

For our texture model, we further assume that the noisy image $y$ also follows the distribution given by \eqref{eq:exponential-family}, with a $\lambda$ that depends both on the underlying texture class of $x$ and on the noise level $\sigma$. This assumption is conceptually reasonable if we think that a noisy texture is still a texture, but the distribution in \eqref{eq:exponential-family} is generally not closed under additive Gaussian noise, so it is best considered as an approximation. Given this, we can express the score of the maximum entropy density in \eqref{eq:exponential-family} as:
\begin{equation}\label{eq:y-score}
    \nabla_y \log p(y) = -\nabla_y [\lambda(x,\sigma)^T  f(y)] .
\end{equation}
Note that $Z(\lambda)$ disappears -- one of the primary motivations for ``score-matching'' methods \citep{hyvarinen2005estimation}. We parameterize neural networks to compute approximations $f_\theta(y)$ and $\lambda_\phi(x, \sigma)$, and express our $\varepsilon$-predictor in terms of their inner product by combining \eqref{eq:miyasawa} and \eqref{eq:y-score}:
\begin{equation}
    \hat{\varepsilon}_{\theta, \phi}(y; x, \sigma) = \frac{\sigma}{\sqrt{\sigma^2 + 1}} \nabla_y [\lambda_\phi(x, \sigma)^T  f_\theta(y)] .
\end{equation}
Note that the dependency of $\hat{\varepsilon}$ on $x$ is only through the low-dimensional vector $\lambda_\phi(x,\sigma)$ and does not make the denoising problem trivial.

In practice, we find that the model performs better when its output is trained to estimate a combination of signal and noise (so-called  $v$-prediction \citep{salimans2022progressive}):
\begin{equation}\label{eq:v-prediction}
    v = \frac{1}{\sqrt{\sigma^2+1}} \varepsilon - \frac{\sigma}{\sqrt{\sigma^2+1}} x .
\end{equation}
Now the effective $\varepsilon$-predictor becomes:
\begin{equation}\label{eq:eps-hat}
    \hat{\varepsilon}_{\theta, \phi}(y; x, \sigma) = \frac{\sigma}{\sqrt{\sigma^2+1}} y - \frac{1}{\sqrt{\sigma^2+1}} \nabla_y [\lambda_\phi(x, \sigma)^T  f_\theta(y)] .
\end{equation}
This reparameterization from $\varepsilon$-prediction to $v$-prediction is without loss of generality if the span of $f(y)$ includes the squared norm $y^T y$ (i.e. there exists a $\lambda$ such that $\nabla_y [\lambda^T  f(y)] = y$), since the difference can be absorbed into $\lambda$. With this form of $\hat{\varepsilon}_{\theta, \phi}(y; x, \sigma)$, we optimize:
\begin{equation}
    \min_{\theta, \phi} \mathbb{E}_{x,\sigma,\varepsilon} \left\Vert \hat{\varepsilon}_{\theta, \phi}(y; x, \sigma) - \varepsilon \right\Vert ^2 .
\end{equation}

This is a minimum mean squared error (MMSE) denoising objective (or equivalently, $\varepsilon$-predictor).  When optimized over a wide range of noise levels $\sigma$, it provides a substrate for a generative diffusion model.

\begin{figure}
    \centering
    \includegraphics[width=1.0\linewidth]{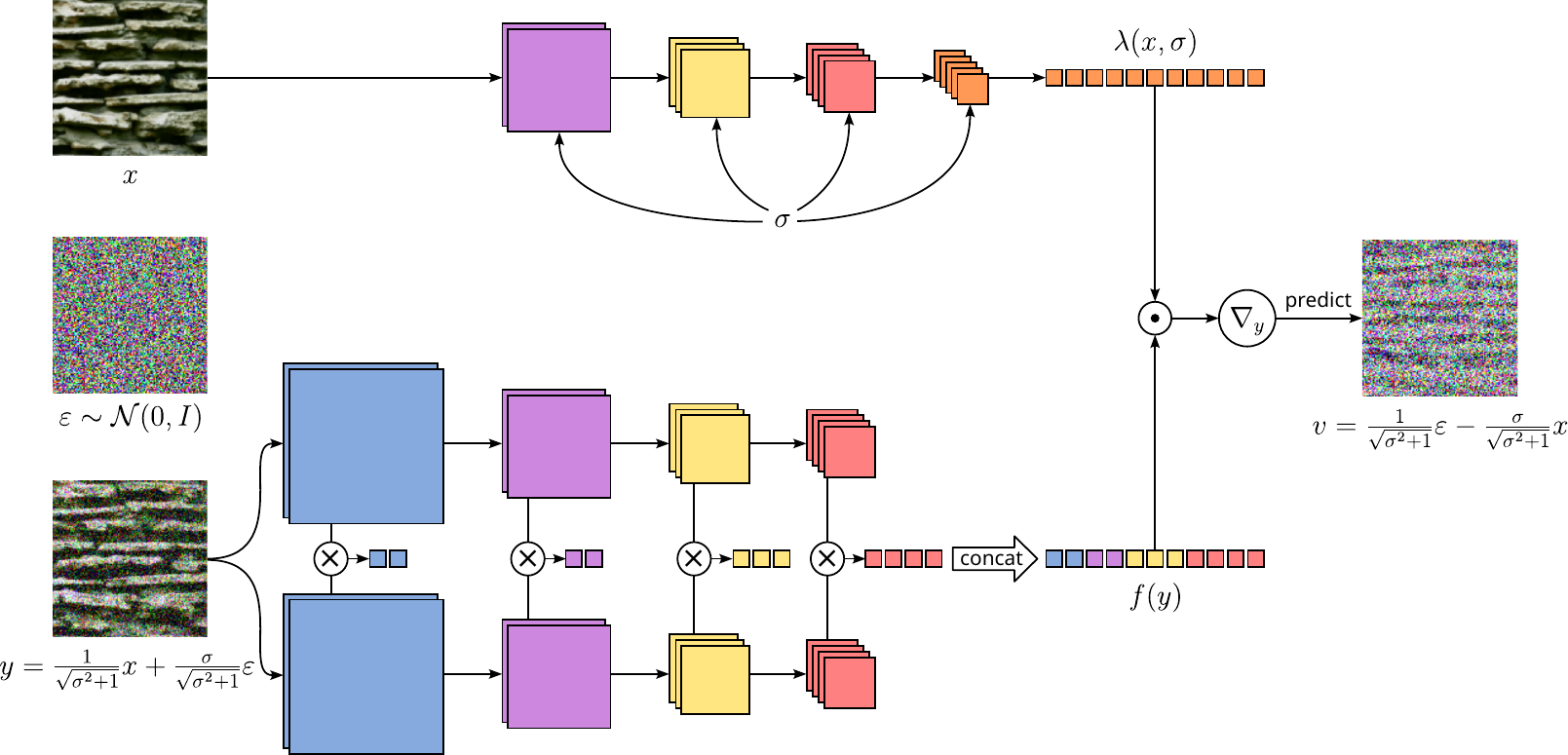}
    \caption{Model diagram. The top and bottom networks compute vectors $\lambda(x,\sigma)$ and $f(y)$, respectively. 
    Input and output images $\{x,\varepsilon,y,v\}$ are 128x128 in resolution. Blue, purple, yellow, red, orange colors represent channels with spatial resolutions 128x128, 64x64, 32x32, 16x16, 8x8, respectively (not drawn to scale). 
    Top network is based on a ConvNeXt-T model \citep{liu2022convnet}, and the $\lambda(x,\sigma)$ vector consists of spatial averages of channels at the last stage of the network.
    Bottom network is constructed from a twin pair of UNet encoders \citep{rombach2022high}, initialized and trained independently.  The $f(y)$ vector consists of the inner products of corresponding channels in the two UNet encoders (for the final layer of each scale, just before downsampling).
    The gradient of the inner product of the two vectors provides an estimate of $v$ (see text).
    }
    \label{fig:diagram}
\end{figure}

\subsection{Dataset}\label{subsec:dataset}

There are several existing texture datasets (e.g., DTD \citep{cimpoi2014describing}, ALOT \citep{burghouts2009material}, OpenSurfaces \citep{bell2013opensurfaces}), but these are generally not large enough to train an expressive neural network from scratch (DTD: 5640 images of 47 classes; ALOT: $\sim$27500 images of 250 classes; OpenSurfaces: 27796 texture regions contained in bigger images). In addition, many images in these dataset are substantially spatially inhomogeneous in luminance, color, orientation, scale and other features, due to perspective, lighting and inherently inhomogeneous structure (e.g. a single large vein on a leaf), but our model (as most texture models) assumes spatial homogeneity. Significantly larger datasets all consist of generic (mostly inhomogeneous) natural images, so we need to select a homogeneous subset for training purposes.

To this end, we crop the natural images in the ImageNet21K dataset \citep{deng2009imagenet} into non-overlapping 128x128 images patches, and then select the most ``homogeneous'' patches using a simple criterion. We first decompose each patch using the steerable pyramid \citep{simoncelli1995steerable} (implemented by \citet{duong2023plenoptic}), a linear transform that partitions the image into feature maps at different orientations and scales. We use a pyramid with 3 scales and 4 orientations, resulting in 4 oriented feature maps with 64x64, 32x32 and 16x16 resolutions, as well as a 128x128 map of the high-pass residual and a 16x16 map of the low-pass residual. We square all 14 feature maps (except for the low-pass residual), and recursively blur and downsample them to 16x16. For each, we compute a homogeneity score - the standard deviation divided by the mean - and sum these over all maps. We chose the one million homogeneous patches with lowest scores as suitable for training our network (see examples in Figure \ref{fig:dataset} in the Appendix).

We do not need a large number of texture images for testing, so we use the Describable Textures Dataset (DTD) \citep{cimpoi2014describing} as the test set. This avoids overlap with the training images (patches drawn from ImageNet), and provides a convenient benchmark on a public database for future comparisons. For preprocessing, we resize each DTD image so that its shorter side has length 128, and then center crop it to 128x128. We use all 5640 DTD images for quantitative evaluations. For visualization, we select 12 images based on the same homogeneity score that we used to generate the training set. Specifically, we compute homogeneity scores for all DTD images, and find that 599 (out of 5640) are above the threshold for training images. We divide these into 12 subsets (each containing 50) and select one image from each subset to ensure the quality and diversity among the final 12 images that are shown in Figures \ref{fig:resample} and \ref{fig:resample2}. We also choose 6 inhomogeneous images (which we do not consider as proper textures) to demonstrate the out-of-distribution behavior of the model (Figure \ref{fig:resample-inhomogeneous}).

\subsection{Network architecture and training}

To construct $\lambda_\phi(x, \sigma)$, we use a ConvNeXt-T model \citep{liu2022convnet} and introduce dependency on $\sigma$ via a sinusoidal embedding and per-channel FiLM modulation \citep{perez2018film} at the end of each convolution block. We replace the stride of the stem convolution from 4 to 2, so that the network can operate on a 128x128 input. See Figure \ref{fig:diagram} for a diagram.

To construct $f_\theta(y)$, we take the first half of a UNet architecture \citep{rombach2022high} and treat the channel activation at the end of each scale (those that would enter the second half of the UNet as skip connections) as the network output. We use two such networks of identical architecture, but whose parameters are initialized and trained independently. We feed them with the same noisy image $y$ and compute inner products between their corresponding output channels, producing one scalar value per channel.  The network output, $f_\theta(y)$, corresponds to the vector containing these scalars.

Here we provide justification for this non-standard ``twin half-UNet'' architecture of the $f$ network. Our initial attempt was to use an identical ConvNeXt as the $\lambda$ network (without $\sigma$ conditioning), but this completely failed on texture generation. We reasoned that we needed to build a network whose {\em gradient} resembles a UNet, which is known to perform well in denoising. Specifically, the network needs to contain input-to-output paths on different spatial scales, such that features of any spatial scale can be computed without going through coarser scales (in a UNet this is enabled by the skip connections). To satisfy this, we tried using the first half of a UNet architecture and computing output by taking the spatial average of each scale, which resulted in a dramatic improvement. We further improved performance by using two identical half-UNet architectures and taking the inner product between corresponding channels along the spatial dimensions.

We set the number of statistics to $d=512$ (increasing this to 1024 did not noticeably increase the performance). We choose a log-normal training distribution for $\sigma$: $\log \sigma \sim \mathcal{N}(0.0, 0.6^2)$. This follows \citet{karras2022elucidating}, except that the mean is larger and the variance is smaller, which we find to work better for our purpose. This distribution is also close to the ``inverse schedule'' used in \citet{hudson2024soda}. We use our curated texture dataset (described in \ref{subsec:dataset}), normalize the images to the range [-1, 1], and use random crop augmentation. We train our network for 10 epochs with batch size 64, using Adam optimizer \citep{kingma2014adam} in PyTorch \citep{ansel2024pytorch}. The training takes 25 hours on four NVIDIA A100 GPUs.

\subsection{Image representation and generative sampling}
Under the maximum entropy framework, both $\lambda(x, \sigma)$ and $f(x)$ can be considered as the ``representation'' of a texture image $x$. Note that given $x$, the former still explicitly depends on $\sigma$. Also note that although the function $f$ takes the noisy image $y$ as its input during training, here we treat it as a representation model for the clean image $x$. Since $f(x)$ is computed on a homogeneous image as a spatial average of convolution channels, the value of $f(x)$ is almost always very close to its expectation $\mu$, by the law of large numbers (more formally, by the equivalence of ensembles or concentration of measure \citep{talagrand1996new}). Therefore, we can assume the approximation $\mu \approx f(x)$. Furthermore, since $\mu$ and $\lambda$ are uniquely determined by each other in the maximum entropy formalism (Appendix \ref{app:maxent}), the two representations $\lambda(x, \sigma)$ and $f(x)$ are approximately equivalent. We can generate new texture images from either of these two representations, by using one of the following two methods. Appendix \ref{app:sampling-relationship} explains how these two methods are mathematically related.

\paragraph{Statistics matching}
Given a reference texture $x_0$, we measure its statistics $\mu_0 = f(x_0)$ and solve the following optimization problem:
\begin{equation}
    \min_x \Vert f(x) - \mu_0 \Vert^2
\end{equation}
We initialize $x$ with i.i.d uniform distribution in the allowed range [-1, 1], and optimize with Adam optimizer for 5000 steps. This method 
is used in a number of existing texture models, including \citet{gatys2015texture}, against which we compare our results.

\paragraph{Diffusion sampling}
Given a schedule of $\sigma_t$ and the corresponding $\lambda_t = \lambda(x_0, \sigma_t)$ for a reference texture $x_0$, we perform the standard DDPM reverse diffusion procedure \citep{ho2020denoising}:
\begin{equation}\label{eq:ddpm}
    x_{t-1} = \frac{1}{\sqrt{\alpha_t}} \left( x_t - \frac{1 - \alpha_t}{\sqrt{1 - \bar{\alpha}_t}} \hat{\varepsilon}(x_t; \lambda_t) \right) + \sqrt{\frac{1 - \bar{\alpha}_{t-1}}{1 - \bar{\alpha}_t} (1 - \alpha_t)} \eta ,
\end{equation}
where $\bar{\alpha}_t = 1/(1+\sigma_t^2)$ and $\alpha_t = \bar{\alpha}_t  / \bar{\alpha}_{t-1} $, and $\eta$ is an identity-covariance Gaussian noise. We use a schedule in which $\sigma_t$ goes from 10.0 to 0.1 in 1000 steps, uniformly spaced on a log scale. 

\subsection{Baseline models}\label{subsec:gatys}
We compare textures with our model to those of several existing \emph{parametric} texture models. There are other methods for texture synthesis, such as text-based methods \citep{wang2024infinite,guerrero2024texsliders}, but these are not directly comparable to our model.

\paragraph{Portilla-Simoncelli model}  \citet{portilla2000parametric} developed a parametric texture model inspired by biological vision, based on a set of hand-selected statistics measured in a multi-scale image representation (a ``steerable pyramid'') \citep{simoncelli1995steerable}. Samples were drawn by matching statistics using a custom gradient-based optimization procedure.  The original model only considered grayscale images, but a publicly available extensions provides an extension to color images\footnote{\url{https://www.cns.nyu.edu/~lcv/texture/}}, which we use for comparison. The color model is based on 6,495 statistics, but about half of these are redundant (due to the symmetry of covariance matrices), yielding $\sim$3k unique statistics.
\paragraph{$\mu$NCA model}
\citet{mordvintsev2021mu} proposed a texture model based on a cellular automata with a small number of free parameters. For each given texture, these parameters need to be trained so that the synthesized image after several iterations is close to the original, according to their hand-designed metric. We use the best performing version of their model, which has 588 parameters, and follow their official implementation\footnote{\url{https://github.com/google-research/self-organising-systems}}.
\paragraph{Gatys model} 
The model introduced in \citet{gatys2015texture} is currently considered the state-of-the-art for parametric texture representation and generation. The statistics it uses are the spatial mean and cross-channel covariance of the hidden layers of the VGG19 network \citep{simonyan2014very} that is pretrained on ImageNet classification. Since the dimensionality of the covariance is proportional to the square of the number of channels in a hidden layer, the Gatys model consists of a very large number of statistics (the standard version has 176,640). New textures are generated via statistics matching. We find that high-quality synthesis requires initialization of $x$ with a small noise amplitude (we use a uniform distribution over 1\% of the allowed range) since the model otherwise produces high-frequency noisy artifacts in the generated texture images (see \ref{subsec:competition}).

\section{Results}

\subsection{Texture resampling}

\begin{figure}[h]
    \centering
    \begin{minipage}[t]{0.03\linewidth}
    \vspace{30pt}
    (a) \\[52pt]
    (b) \\[52pt]
    (c) \\[52pt]
    (d) \\[52pt]
    (e) \\[52pt]
    (f)
    \end{minipage}
    \begin{minipage}[t]{0.96\linewidth}
    \vspace{0pt}
    \includegraphics[width=1.0\linewidth]{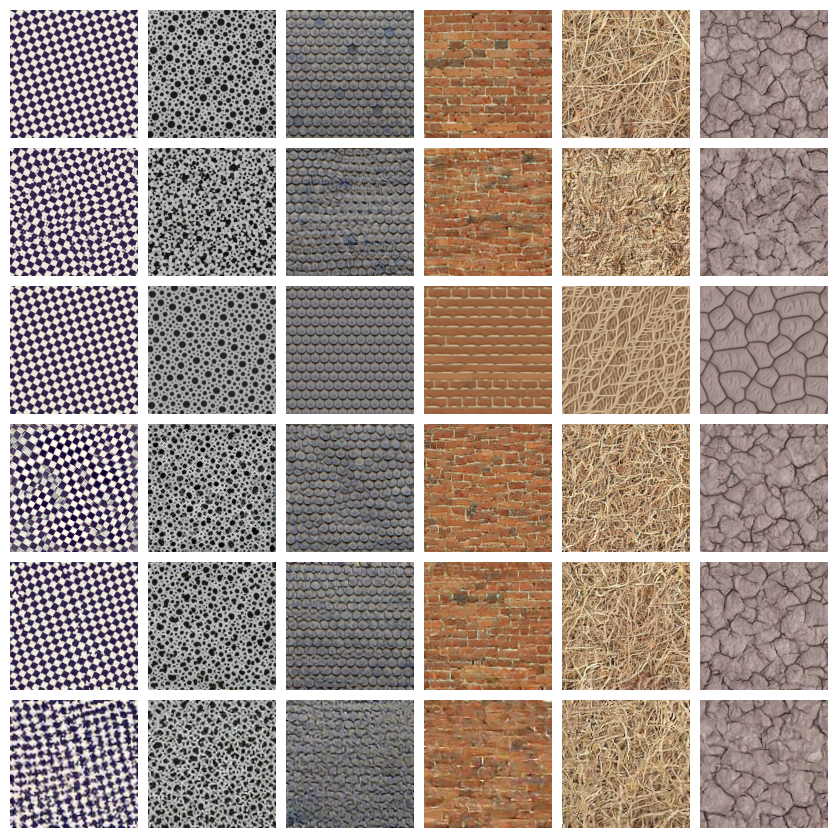}
    \end{minipage}
    
    \caption{Resampling examples.
    \textbf{(a)} Original texture images (from DTD dataset, sorted in decreasing homogeneity).  Remaining columns show samples generated by 
    \textbf{(b)} Portilla-Simoncelli model;
    \textbf{(c)} $\mu$NCA model;
    \textbf{(d)} Gatys model;
    \textbf{(e)} statistics matching for our model;
    \textbf{(f)} diffusion sampling for our model. }
    \label{fig:resample}
\end{figure}

Given an original texture image $x_0$, we can synthesize a new sample (i.e. generate a new texture image in the same class as $x_0$) by matching statistics $\mu_0 = f(x_0)$, or by diffusion sampling conditioned on $\lambda_0 = \lambda(x_0, \sigma)$. Over a set of test images, we find that statistics matching generates better quality images than diffusion sampling (Figure \ref{fig:resample}; additional examples in Figure \ref{fig:resample2}; examples for inhomogeneous images in Figure \ref{fig:resample-inhomogeneous}). For comparison, we find that our model (with statistics matching) generates texture images whose visual quality is similar to or better than those generated with the Gatys model, and is significantly better than the Portilla-Simoncelli model. The $\mu$NCA model performs very well on textures that are largely periodic, but produces overly smooth or stylized samples for less regular textures. Note that our model uses significantly fewer (512) parameters than the other models, especially than the Gatys model (176,640 parameters, which exceeds the number of pixels in the 128x128 training images). It is intriguing that the two sampling methods for our model lead to different resampling quality, since the two representations $\mu$ and $\lambda$ are related by a one-to-one correspondence (bijection).  In addition, as we explain in Appendix \ref{app:sampling-relationship}, the two sampling methods are also mathematically related.

\subsection{Perceptual quality metric}

\begin{figure}
    \centering
    \includegraphics[width=0.49\linewidth]{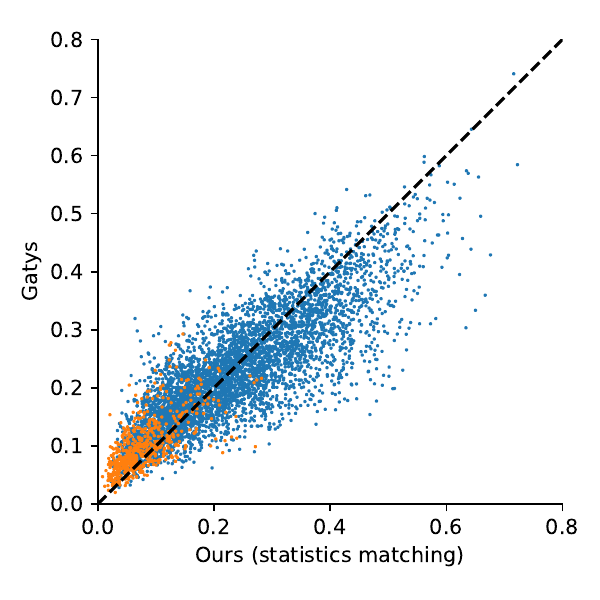}
    \hfill
    \includegraphics[width=0.49\linewidth]{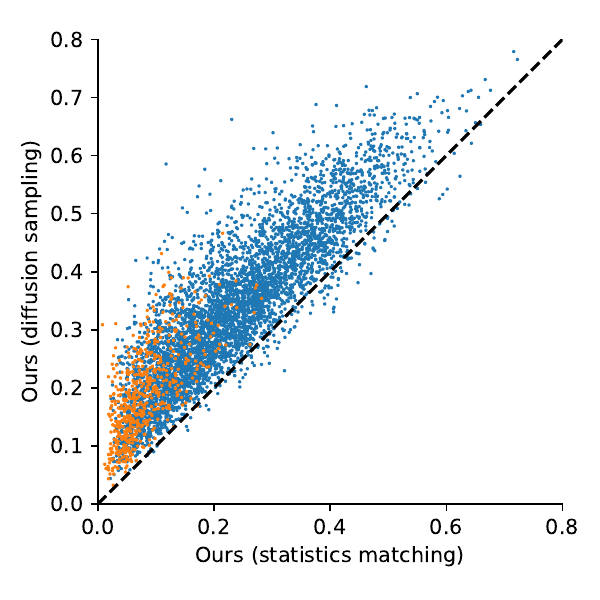}
    \caption{Perceptual quality evaluated using DreamSim (smaller is better). Each point corresponds to one image in the DTD dataset. Color labeling designates homogeneity of each image, compared with the threshold used to select the training set (\textcolor{C1}{orange: more homogeneous}, \textcolor{C0}{blue: less homogeneous}).
    Left: our model with statistics matching v.s.\ the Gatys model; right: our model with statistics matching v.s.\ with diffusion sampling. In both panels, points above the dashed identity line correspond to images on which our model with statistics matching performs better.}
    \label{fig:dreamsim}
\end{figure}

To quantitatively compare the sample quality of different models, we compute the DreamSim score \citep{fu2023dreamsim} between original and resampled images. We use all 5,640 images in the DTD dataset \citep{cimpoi2014describing} as originals, and generate 10 resampled images per original, for our model (with statistics sampling) and the Gatys model. We find that scores vary substantially across the images in the DTD dataset, and are dependent on the homogeneity score of the original (as defined in \ref{subsec:dataset}). Out of the 5,640 images, 599 images are more homogeneous than the threshold that we used for the selection of the training set (which resulted in 1 million training patches). The rest (5,041 images) are less homogeneous than the threshold. As the left panel of Figure \ref{fig:dreamsim} shows, both models perform better (have smaller scores) on the more homogeneous images (orange dots) than the less homogeneous images (blue dots). Importantly, the two models perform similarly on the less homogeneous images, but our model (with statistical matching) outperforms the Gatys model on a majority of the more homogeneous images (Table \ref{tab:dreamsim}). The right panel of Figure \ref{fig:dreamsim} shows that for our model, statistics matching outperforms diffusion sampling on almost all DTD images.

The DreamSim score (like other perceptual quality scores such as DISTS \citep{ding2020image} and LPIPS \citep{zhang2018unreasonable}) is computed on pairs of individual images (one original and one sample), and therefore cannot quantify sample diversity. Visually, when we repeat the sampling procedures with different random seeds, all methods seem to generate different samples (Figures \ref{fig:resample-repeat-1} and \ref{fig:resample-repeat-2} in the Appendix). Although this shows that the models do not always generate the same image (which could happen if the model memorizes the training set), it does not reveal more subtle aspects of diversity. To quantify diversity, we computed the FID scores (Appendix \ref{subsec:fid}), which estimates a distance between the distributions of latent representations of true and generated images, providing a measure of both quality and diversity.

\begin{table}
  \caption{Comparison of our model (with statistics matching) and Gatys model using DreamSim score. For each original image, Welch's $t$-test is performed using 10 samples for each model, and $p>0.05$ is deemed a tie.}
  \label{tab:dreamsim}
  \centering
  \begin{tabular}{lrrrr}
    \toprule
    & Total & Ours better & Gatys better & Tie \\
    \midrule
    All DTD images & 5640 & 2398 & 2112 & 1130 \\
    \textcolor{C1}{More homogeneous} & 599 & 411 & 118 & 70 \\
    \textcolor{C0}{Less homogeneous} & 5041 & 1987 & 1994 & 1060 \\
    \bottomrule
  \end{tabular}
\end{table}

\subsection{Interpolation}

\begin{figure}
    \centering
    \begin{minipage}[t]{0.03\linewidth}
    \vspace{30pt}
    (a) \\[52pt]
    (b) \\[52pt]
    (c) \\[52pt]
    (d) \\[52pt]
    (e)
    \end{minipage}
    \begin{minipage}[t]{0.96\linewidth}
    \vspace{0pt}
    \includegraphics[width=1.0\linewidth]{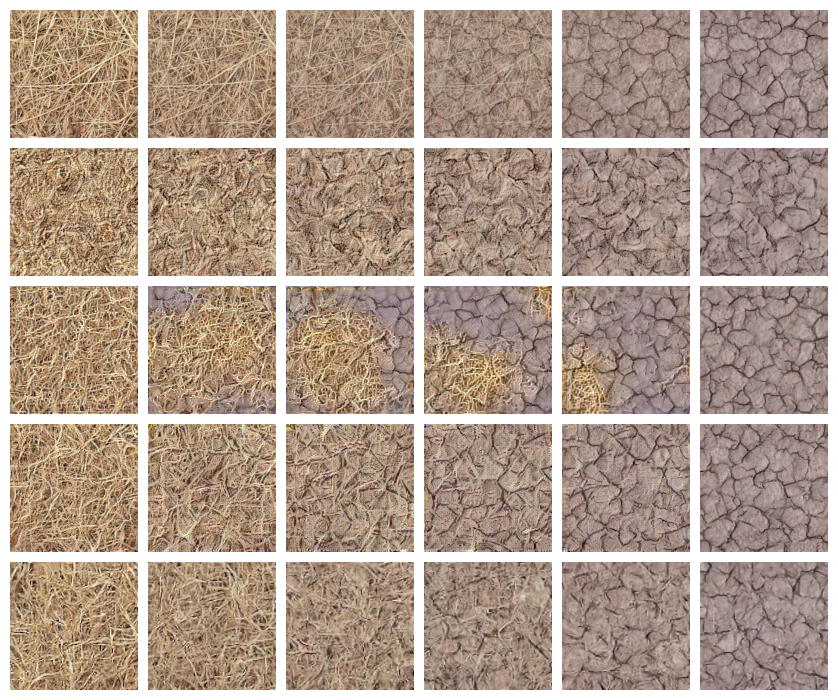}
    \end{minipage}
    
    \caption{Interpolation between the 5th and 6th images in Figure \ref{fig:resample}. \textbf{(a)} Fading in pixel values. \textbf{(b)} Portilla-Simoncelli model. \textbf{(c)} Gatys model. \textbf{(d)} Statistics matching for our model. \textbf{(e)} Diffusion sampling for our model.}
    \label{fig:interpolate}
\end{figure}

We examine the structure and generalization capabilities of our model's representation space by generating texture samples whose parameters lie between those of two existing textures $x_a$ and $x_b$. Linear interpolation can be performed in $\mu$ or $\lambda$ space, using statistics matching or diffusion sampling, respectively:
\begin{equation}
    \mu_m = \frac{M-m}{M} f(x_a) + \frac{m}{M} f(x_b), \ \ {\rm or} \ \ \lambda_m(\sigma) = \frac{M-m}{M} \lambda(x_a, \sigma) + \frac{m}{M} \lambda(x_b, \sigma), \quad m=0,\cdots,M
\end{equation}
Note that interpolation in these two spaces is generally not  equivalent: $\lambda$ and $\mu$ are related by a highly nonlinear bijection, so a straight line in one space is likely to be curved in the other. For comparison, we perform the same task using other parametric texture models (Portilla-Simoncelli, $\mu$NCA, Gatys) by linearly interpolating (going in straight lines) in their respective representation spaces.

Figure \ref{fig:interpolate} provides a set of example interpolations (additional examples are provided in Appendix Figures \ref{fig:interpolate2} and \ref{fig:interpolate3}). Interpolation results for $\mu$NCA are not shown because the cellular automata dynamics becomes unstable with the interpolated parameters.
Interpolation in the pixel domain (i.e., ``fading'' between images) produces ``double exposure'' images [row (a)].
Samples generated to match interpolated statistics in the Gatys model [row (c)] are patchwise spatial mixtures of the two endpoint textures, rather than homogeneous textures with features that lie between those of the endpoint textures. If we consider the valid support set of these models to be restricted to homogeneous textures, then these examples provide evidence that the representation space of the Gatys model is not convex.
In contrast, interpolated samples from the Portilla-Simoncelli model and from our model (interpolated in $\mu$ or $\lambda$) are seen to be homogeneous, with features that smoothly transition from those of the two endpoint textures.\footnote{Note that the original Portilla-Simoncelli paper \citep{portilla2000parametric} reported that samples for interpolated parameters appeared as patchwise spatial mixtures (like the Gatys model), but those examples were not computed by linearly interpolating the model parameters.  Instead, the parameters were transformed to expected values (e.g., correlations are replaced by unnormalized and uncentered second moments) and linear interpolation was performed there. The rationale was to guarantee the existence of images that are consistent with the interpolated parameters. Linear interpolation on these nonlinearly transformed parameters is not equivalent to linear interpolation on the original parameters, as was done for the examples in row (b) of Figures \ref{fig:interpolate}, \ref{fig:interpolate2} and \ref{fig:interpolate3}.} 
For $\mu$-space interpolation via statistics matching, samples sometimes contain sporadic artifacts, especially at their boundaries (this is present in Figure \ref{fig:interpolate} but more visible in Figure \ref{fig:interpolate3}).
For interpolation of $\lambda$-space via diffusion, such artifacts are not present, although the quality is often a bit worse than that of statistics matching (as is shown in Figure \ref{fig:resample}).

\section{Discussion}

We have developed a framework for learning a maximum entropy density model from data. Specifically, given a training set of texture images, we learn an exponential family model that is defined in terms of a set of statistics, such that each image corresponds to a sample from the maximum entropy density constrained to match those statistics.
We use a two-network architecture (one computing the statistics, and other the weights) in which their combination is an inner product, as dictated by the maximum entropy construction: the log probability is a weighted sum of the statistics. 
The full model is optimized for denoising performance, as is done for generative diffusion models. 

We find that images generated by our model trained on a large dataset of textures
are visually superior to the state-of-the-art Gatys model, while relying on a far smaller parameter set (512 vs.\ 176,640).
Our model formulation enables two generative sampling methods: statistics matching and diffusion sampling, and we find that the former generates higher quality samples. We suspect that this may be related to the hypothesis in our mathematical framework that the exponential family distribution needs to be closed under additive Gaussian noise, which is generally not exactly true. Deviation from this hypothesis may hurt diffusion sampling more than statistics matching, since diffusion sampling depends on the accurate estimation of scores at every noise level, while statistics matching only aims to match the statistics for the \emph{clean} image and is adaptive during the optimization process (see Appendix \ref{app:sampling-relationship}).

We also show that a straight trajectory between the representations ($\mu$ or $\lambda$) of two given textures generates a visually smooth sequence of interpolated textures, suggesting that the representation space of our model has convexity properties that are desirable for texture manipulation (e.g., style transfer). We are also interested in discovering interpretable or semantically meaningful directions in the representation space (as in \citet{hudson2024soda}), either with unsupervised dimensionality reduction methods, or using category labels in datasets such as DTD \citep{cimpoi2014describing}.

Our texture model can serve as a scientific tool for generating stimuli for neuroscience and psychology experiments \citep{victor2017textures}. Several existing papers \citep{freeman2013functional,lieber2025responses} are based on the texture model in \citet{portilla2000parametric}, which is relatively low-dimensional (710 statistics for grayscale images, $\sim$3k for color images) and more interpretable by construction, but is worse in visual quality than those based on deep neural networks. \citet{jagadeesh2022texture} uses the Gatys model to generate stimuli for experiments, which has a very large dimensionality ($\sim$177k statistics). With a much smaller size (512 statistics), our model offers a decisive improvement in both model size and visual quality, opening the possibility of characterizing neural responses in the representation space.

Finally, our method is general.  As with generative diffusion models, it relies on the use of networks with inductive biases that are well-matched to the target density, and can be applied to any data modality that can be appropriately described by a maximum entropy model, such as homogeneous temporal segments of sounds \citep{mcdermott2011sound}, spatiotemporal patches of videos \citep{xie2019learning}, or spiking data measured from biological neurons \citep{schneidman2006weak,mayzel2024homeostatic,lynn2025exact}. Learning statistics directly from the data could lead to improved models for these modalities, as it has for visual textures.


\newpage
\begin{ack}
    We would like to thank Florentin Guth for suggesting the twin half-UNet architecture of the $f$ network (bottom half of Figure \ref{fig:diagram}).

    This work was supported by the Flatiron Institute at the Simons Foundation.
\end{ack}

\bibliography{biblio}


\newpage
\appendix
\section{Maximum entropy probability distribution}
\label{app:maxent}

\subsection{Derivation of the exponential family}
We rewrite \eqref{eq:maxent-problem} as:
\begin{equation}
    \min_p \int p(x) \log p(x) dx \quad \text{s.t.} \ \int p(x) dx = 1, \ \int p(x) f_k(x) dx = \mu_k,\quad k=1,2,\cdots,d
\end{equation}
Combining the objective and constraints using the method of Lagrange multipliers gives a combined objective:
\begin{equation}
    \int p(x) \log p(x) dx + \nu \left( \int p(x) dx - 1 \right) + \sum_{k=1}^d \lambda_k \left( \int p(x) f_k(x) dx - \mu_k \right)
\end{equation}
Taking the functional derivative with respect to $p$ and setting it to zero gives:
\begin{align}
    & \log p(x) + 1 + \nu + \sum_{k=1}^d \lambda_k f_k(x) = 0 \\
    \Rightarrow \ & p(x) = \exp \left( -1-\nu - \sum_{k=1}^d \lambda_k f_k(x) \right)
\end{align}
Since one of the constraints is that $p$ is normalized, we can eliminate $\nu$:
\begin{equation}
     \exp (1 + \nu) = \int \exp \left( - \sum_{k=1}^d \lambda_k f_k(x) \right) dx =: Z(\lambda)
\end{equation}
Then we arrive at an exponential family distribution \eqref{eq:exponential-family}:
\begin{equation}
    p(x) = \frac{1}{Z(\lambda)} \exp \left( - \sum_{k=1}^d \lambda_k f_k(x) \right)
\end{equation}
The rest of the Lagrange multipliers $\{\lambda_1, \lambda_2, \cdots, \lambda_d\}$ are chosen to satisfy the other constraints, and are thus a function of the statistical measurements $\{\mu_1, \mu_2, \cdots \mu_d\}$ . In most cases, these relationships cannot be expressed in closed form (the Gaussian distribution is an exception, as we show below).

\subsection{Example: the Gaussian distribution}
Here, we consider the simplest example of  maximum-entropy density model. Suppose $x$ is a scalar random variable, and one chooses to constrain the density using two statistics $f(x) = [x, x^2]$ (i.e., the first and second moments). Then the maximum entropy distribution is:
\begin{equation}
    p_\lambda(x) = \frac{1}{Z(\lambda)} \exp( -\lambda_1 x - \lambda_2 x^2)
\end{equation}
The parameters $\lambda_1$ and $\lambda_2$ must be chosen to satisfy the constraints $\mathbb{E}(x) = \mu_1$ and $\mathbb{E}(x^2) = \mu_2$. In general, $\lambda = [\lambda_1, \lambda_2]$ cannot be written analytically as a function of $\mu = [\mu_1, \mu_2]$, but in this case we see that the density is an exponential of a quadratic form, and thus a Gaussian distribution. So we can solve directly, giving:
\begin{equation}
    \lambda_1 = -\frac{\mu_1}{\mu_2 - \mu_1^2}, \qquad \lambda_2 = \frac{1}{2(\mu_2 - \mu_1^2)}
\end{equation}
Note that $\mu_2 - \mu_1^2$ is the variance (second moment minus squared first moment). The mapping between $\mu$ and $\lambda$ is one-to-one, and either can serve as a parametrization of the family of Gaussian distributions.

\subsection{One-to-one correspondence between $\lambda$ and $\mu$}
Define the set of $\lambda$ that leads to a normalizable distribution:
\begin{equation}
    \Lambda = \{\lambda \in \mathbb{R}^d: Z(\lambda) < +\infty\}
\end{equation}
We also know that the set of $\mu$ that is achievable by some $p$ is:
\begin{equation}
    M = \{\mu \in \mathbb{R}^d: \mu = \mathbb{E}_{x\sim p} [f(x)], \exists p\} = \text{conv} (\{f(x): x\in\mathbb{R}^n\})
\end{equation}
where conv denotes the convex hull.

It is always true that $\lambda$ uniquely determines $\mu$. When the following two conditions are met:
\begin{itemize}
    \item minimal: there is no $\lambda$ such that $\sum_{i=1}^d \lambda_k f_k(x) = \text{constant}$.
    \item regular: $\Lambda$ is an open set (true for all useful distributions).
\end{itemize}
Then it is also true that any $\mu$ in the interior of $M$ uniquely determines $\lambda$. In other words, we have a one-to-one correspondence (bijection) between $\Lambda$ and $\text{interior}(M)$. \footnote{See {\tt https://www.stat.umn.edu/geyer/5421/notes/expfam.pdf} for a mathematically rigorous treatment.}

\section{Tweedie's formula for a variance-preserving noise model}
\label{app:miyasawa}
Consider the variance-preserving noise process:
\begin{equation}
    y = \frac{1}{\sqrt{\sigma^2 + 1}} x + \frac{\sigma}{\sqrt{\sigma^2 + 1}} \varepsilon
\end{equation}
where $\varepsilon$ is Gaussian noise with idenity covariance.  Then
\begin{equation}
    p(y|x) = \mathcal{N} \left( y; \frac{1}{\sqrt{\sigma^2 + 1}} x,  \frac{\sigma^2}{\sigma^2 + 1} I \right) 
    \propto 
    \exp \left( -\frac{\sigma^2 + 1}{2\sigma^2} \left\Vert y - \frac{1}{\sqrt{\sigma^2 + 1}} x \right\Vert^2 \right)
\end{equation}
Then we compute the score (i.e. the gradient of log probability):
\begin{align}
    \nabla_y \log p(y) &= \frac{1}{p(y)} \nabla_y p(y) \\
    &= \frac{1}{p(y)} \int p(x) \nabla_y p(y|x) dx \\
    &= \frac{1}{p(y)} \int p(x) p(y|x) \left(- \frac{\sigma^2+1}{\sigma^2}\right) \left(y - \frac{1}{\sqrt{\sigma^2 + 1}} x \right) dx \\
    &= -\int \frac{p(x) p(y|x)}{p(y)} \left(\frac{\sqrt{\sigma^2 + 1}}{\sigma}\right) \left( \frac{\sqrt{\sigma^2+1}}{\sigma} y - \frac{1}{\sigma} x \right) dx \\
    &= -\frac{\sqrt{\sigma^2 + 1}}{\sigma} \int p(x|y)  \varepsilon (x, y) dx \\
    &= -\frac{\sqrt{\sigma^2 + 1}}{\sigma} \mathbb{E}[\varepsilon|y]
\end{align}

\section{Relationship between statistics matching and diffusion sampling}
\label{app:sampling-relationship}

In statistics matching of $f: \mathbb{R}^n \to \mathbb{R}^d$ for a given base image $x_0 \in \mathbb{R}^n$, we optimize:
\begin{equation}
    \arg\min_x \Vert f(x) - f(x_0) \Vert^2
\end{equation}
If we use standard gradient descent (instead of Adam, which was used to generate samples in all experiments), the update step is:
\begin{equation}
    x \leftarrow x - \beta J_f(x)^T [f(x) - f(x_0)]
\end{equation}
where $\beta>0$ is the learning rate, and $J_f: \mathbb{R}^n \to \mathbb{R}^{d \times n}$ is the Jacobian matrix of $f$.

In the DDPM procedure of \eqref{eq:ddpm}, an update step in diffusion sampling moves the image $x$ in a direction that is a linear combination of $x$ and the $\varepsilon$-predictor $\hat{\varepsilon}(x; \lambda)$, where $\lambda = \lambda(x_0, \sigma_t)$ at each time step $t$ is determined by the base image $x_0$ and noise schedule $\sigma_t$. According to \eqref{eq:eps-hat}, the $\varepsilon$-predictor is a linear combination of $x$ and the output of the network, which is:
\begin{equation}
    \nabla_x [\lambda^T  f(x)] = J_f(x)^T \lambda(x_0, \sigma_t)
\end{equation}
Together, these mean that the update step for diffusion sampling moves $x$ in the subspace spanned by $x$ and the rows of $ J_f(x)$. If we assume $x \in \text{span} \{\text{rows of } J_f(x)\}$ (the same assumption we made to justify the reparameterization of $v$-prediction), then we can rearrange terms in the update rule, such that there is a $\tilde{\lambda}$ (determined by $x_0$ and the noise schedule $\sigma_t$ for each time step $t$) such that the update rule reads:
\begin{equation}
    x \leftarrow x - J_f(x)^T \tilde{\lambda}(x_0, \sigma_t) + \text{noise}
\end{equation}

Both update rules move $x$ in a direction that lies in the subspace spanned by the rows of $J_f(x)$. Apart from the additive noise, the fundamental difference between the two rules is that the vector-Jacobian product with $J_f(x)$ (which linearly combines its rows) is done with $f(x) - f(x_0)$ in statistics matching, and with $\tilde{\lambda}(x_0, \sigma_t)$ in diffusion sampling. The former is adaptive to the value of $x$ during the optimization process, while the latter is predetermined for each step.

\section{Additional results}

\subsection{Competitive adversarial comparison}\label{subsec:competition}
In order to more directly compare two generative texture models based on statistical measurements $f$ and $g$,
we let them directly compete to expose each other's flaws, using a method similar to MAD competition \citep{wang2008maximum}. Given a texture image $x_0$, we synthesize an image $x$ that has the same representation as $x_0$ according to one model, but is maximally different according to the other:
\begin{equation}
    \max_x \Vert f(x) - f(x_0) \Vert^2 \qquad \text{s.t. } g(x) = g(x_0) .
\end{equation}
In practice, we optimize:
\begin{equation}
    \max_x \left[ \log \left( \frac{\Vert f(x) - f(x_0) \Vert^2}{\Vert f(x_0) \Vert^2} + \delta \right) -  \log \left( \frac{\Vert g(x) - g(x_0) \Vert^2}{\Vert g(x_0) \Vert^2} + \delta  \right) \right] ,
\end{equation}
and we also optimize the same objective but with $f$ and $g$ switched. In order to stabilize the training and make sure the constraint is met, we find it helpful to run optimization with only the second term (equivalent to statistics matching) for 500 steps before using the full loss function for another 4500 steps. The constraint is adequately enforced when using a small $\delta = 10^{-5}$. To determine which model is superior, we compared the two generated images in terms of their perceptual quality and deviation from the original texture $x_0$.

\begin{figure}
    \centering
    \begin{minipage}[t]{0.03\linewidth}
    \vspace{30pt}
    (a) \\[52pt]
    (b) \\[52pt]
    (c) \\[52pt]
    (d)
    \end{minipage}
    \begin{minipage}[t]{0.96\linewidth}
    \vspace{0pt}
    \includegraphics[width=1.0\linewidth]{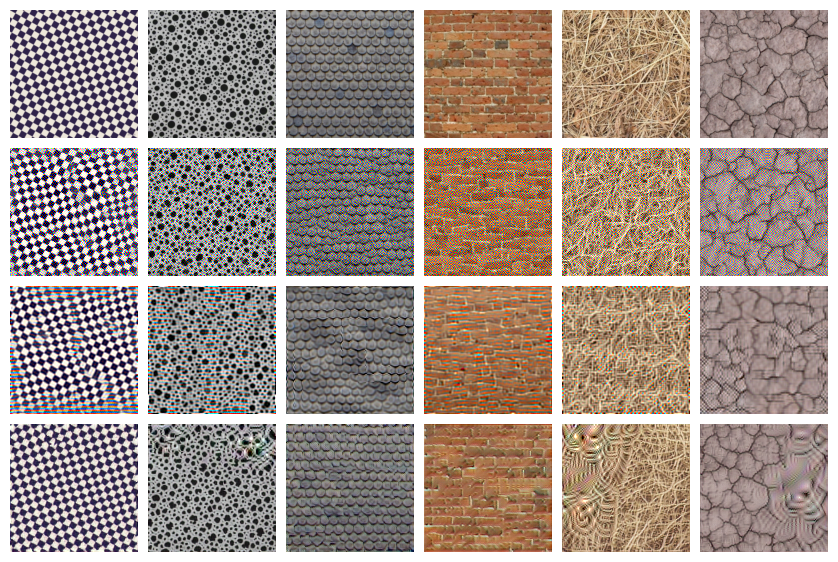}
    \end{minipage}
    
    \caption{Competitive adversarial comparison between the Gatys model and our model.
    \textbf{(a)} Original texture images (as in Figure \ref{fig:resample}).
    \textbf{(b)} Images that are considered the same as (a) according to the Gatys model and maximally different according to our model (zoom in to see the high frequency details).
    \textbf{(c)} Same as (b) but adding a high-pass statistic to the Gatys model.
    \textbf{(d)} Images that are considered the same as (a) according to our model and maximally different according to the Gatys model.}
    \label{fig:competition}
\end{figure}

We apply this method to compare the Gatys model and our own (Figure \ref{fig:competition}; additional examples in Figure \ref{fig:competition2}). Row (b) is dominated by jarring high frequency artifacts, showing that the standard Gatys model does not constrain the high frequency component of the image. To prevent the objective from exclusively exploiting this type of distortion, we add a single high-pass statistic (taken from the steerable pyramid \citep{simoncelli1995steerable}) to the Gatys model. When the Gatys model is complemented with this high-pass statistic, it prevents the high frequency artifacts instead allows the appearance of localized gratings in some textures, and changes in local luminance for mosaic-like textures (row (c)). Our model shows its blind spot in row (d), where it allows for a particular type of artifact made of localized oriented structures for most test textures. 

\subsection{FID score}\label{subsec:fid}
We quantitatively evaluated the quality and diversity of the generated textures using the FID score \citep{heusel2017gans}. To do this, we use several high-resolution texture images \footnote{Downloaded from \url{unsplash.com}} with a large number of repeating units, so that we can generate a large number of crops, all of which are homogeneous textures of the same class. For each image, we compute the representation on 10,000 randomly cropped patches, average over them to obtain one single representation, and resample 10,000 patches based on this representation using the aforementioned methods. The FID score is then calculated against the original 10,000 crops (Table \ref{tab:fid}). For these 8 of the 9 images, the best-performing model is either statistics matching or diffusion sampling. We note that the scores for both our model and the Gatys model are significantly worse than the state-of-the-art values for common data sets (such as ImageNet), but FID scores are not directly comparable across data sets. Moreover, the FID score is constructed using a network trained for object recognition, and is closely tied to the object categories of ImageNet. It has been shown to provide a poor measure of visual quality when used on a model that generates images containing objects outside the categories of ImageNet \citep{kynkaanniemi2022role}. 
There are many alternative metrics \citep{binkowski2018demystifying,kynkaanniemi2019improved,jayasumana2024rethinking}, but they are also based on networks trained on a specific task (in most cases, object recognition), and on corresponding images that are out-of-distribution for our texture model.
In general, we lack a general method for evaluating both quality and diversity of generative models for any data distribution.

\begin{table}
  \caption{FID scores. From top to bottom: statistics matching for the Gatys model, statistics matching for our model, diffusion sampling for our model.}
  \label{tab:fid}
  \centering
  \begin{tabular}{llllllllll}
    \toprule
    & Grass & Pebble & Star & Cloth & Rug & Flower & Marble & Rubber & Glitter \\
    \midrule
    Gatys & 175.03 & \textbf{169.35} & 209.10 & 126.79 & 273.30 & 74.03 & 135.17 & 108.01 & 96.19 \\
    Stat. & \textbf{102.58} & 214.77 & 68.19 & \textbf{41.70} & 105.96 & 80.49 & 137.79 & \textbf{102.46} & \textbf{20.82} \\
    Diff. & 221.30 & 192.94 & \textbf{48.78} & 66.25 & \textbf{55.49} & \textbf{67.14} & \textbf{46.68} & 142.12 & 36.82 \\
    \bottomrule
  \end{tabular}
\end{table}

\subsection{Comparison of perceptual metrics}
Unlike FID, perceptual metrics such as DreamSim \citep{fu2023dreamsim}, LPIPS \citep{zhang2018unreasonable} and DISTS \citep{ding2020image} quantify the distance between two individual images (instead of two distributions), and computing them do not require a large number of crops of the same texture. However, the images generated for computing FID give us a way to compare the perceptual metrics themselves, by testing whether these metrics recover the human perception that the distance between two crops from the original texture is smaller than the distance between one crop and one generated resampled image. To do this, we randomly pair 10,000 crops from the original images with 10,000 generated resampled images and obtain 10,000 scores per metric per generative method (Gatys, ours with statistics matching, ours with diffusion sampling), and we similarly compute the scores between two sets of independent 10,000 crops (Table \ref{tab:perceptual-scores}). We find that DreamSim most consistently rated the distance between two original crops as the smallest. Therefore, we choose to report DreamSim scores in the main text (Figure \ref{fig:dreamsim} and Table \ref{tab:dreamsim}).

\begin{table}
\centering
\caption{DreamSim, LPIPS and DISTS perceptual metrics. Mean and standard deviation are computed over 10,000 samples in each table entry. Smaller number is better (more similar).}
\label{tab:perceptual-scores}
\begin{tabular}{llllll}
\toprule
& & Two crops & Gatys & Stat. matching & Diffusion \\
\midrule
\multirow{9}{*}{DreamSim}
& Grass & \textbf{0.198 ± 0.060} & 0.294 ± 0.042 & 0.273 ± 0.043 & 0.268 ± 0.058 \\
& Pebble & \textbf{0.176 ± 0.046} & 0.224 ± 0.037 & 0.332 ± 0.049 & 0.354 ± 0.063 \\
& Star & \textbf{0.101 ± 0.042} & 0.208 ± 0.029 & 0.143 ± 0.028 & 0.122 ± 0.034 \\
& Cloth & \textbf{0.078 ± 0.034} & 0.131 ± 0.033 & 0.091 ± 0.031 & 0.113 ± 0.032 \\
& Rug & \textbf{0.119 ± 0.037} & 0.232 ± 0.026 & 0.121 ± 0.024 & 0.143 ± 0.031 \\
& Flower & 0.103 ± 0.034 & 0.118 ± 0.030 & \textbf{0.094 ± 0.029} & 0.134 ± 0.033 \\
& Marble & \textbf{0.143 ± 0.034} & 0.211 ± 0.028 & 0.172 ± 0.028 & 0.193 ± 0.035 \\
& Rubber & \textbf{0.134 ± 0.031} & 0.189 ± 0.026 & 0.173 ± 0.028 & 0.200 ± 0.030 \\
& Glitter & 0.118 ± 0.042 & 0.136 ± 0.031 & \textbf{0.115 ± 0.033} & 0.136 ± 0.039 \\
\midrule
\multirow{9}{*}{LPIPS}
& Grass & 0.288 ± 0.022 & \textbf{0.272 ± 0.019} & 0.281 ± 0.019 & 0.284 ± 0.020 \\
& Pebble & 0.436 ± 0.033 & \textbf{0.429 ± 0.029} & 0.441 ± 0.028 & 0.456 ± 0.029 \\
& Star & 0.345 ± 0.046 & 0.337 ± 0.033 & \textbf{0.336 ± 0.036} & \textbf{0.336 ± 0.037} \\
& Cloth & 0.130 ± 0.022 & 0.132 ± 0.017 & \textbf{0.128 ± 0.017} & 0.135 ± 0.021 \\
& Rug & 0.244 ± 0.036 & 0.245 ± 0.023 & \textbf{0.236 ± 0.021} & 0.266 ± 0.023 \\
& Flower & 0.307 ± 0.027 & \textbf{0.299 ± 0.022} & 0.302 ± 0.021 & 0.307 ± 0.022 \\
& Marble & 0.342 ± 0.028 & \textbf{0.325 ± 0.024} & 0.343 ± 0.025 & 0.341 ± 0.027 \\
& Rubber & 0.299 ± 0.023 & \textbf{0.293 ± 0.019} & \textbf{0.293 ± 0.020} & 0.298 ± 0.021 \\
& Glitter & 0.312 ± 0.030 & \textbf{0.301 ± 0.024} & 0.304 ± 0.023 & 0.306 ± 0.023 \\
\midrule
\multirow{9}{*}{DISTS}
& Grass & 0.273 ± 0.021 & \textbf{0.271 ± 0.016} & 0.290 ± 0.018 & 0.314 ± 0.024 \\
& Pebble & \textbf{0.245 ± 0.022} & 0.268 ± 0.017 & 0.277 ± 0.018 & 0.293 ± 0.022 \\
& Star & \textbf{0.239 ± 0.034} & 0.278 ± 0.023 & 0.268 ± 0.022 & 0.251 ± 0.021 \\
& Cloth & 0.207 ± 0.022 & 0.207 ± 0.016 & \textbf{0.205 ± 0.017} & 0.220 ± 0.019 \\
& Rug & 0.254 ± 0.030 & 0.261 ± 0.020 & \textbf{0.247 ± 0.021} & 0.261 ± 0.020 \\
& Flower & 0.219 ± 0.023 & 0.219 ± 0.015 & \textbf{0.217 ± 0.016} & \textbf{0.217 ± 0.015} \\
& Marble & \textbf{0.212 ± 0.015} & 0.216 ± 0.014 & 0.226 ± 0.015 & 0.217 ± 0.014 \\
& Rubber & \textbf{0.246 ± 0.018} & 0.257 ± 0.014 & 0.265 ± 0.015 & 0.295 ± 0.017 \\
& Glitter & 0.233 ± 0.019 & 0.243 ± 0.018 & \textbf{0.229 ± 0.016} & 0.235 ± 0.015 \\
\bottomrule
\end{tabular}
\end{table}

\section{Supplementary figures}

\begin{figure}[h]
    \centering
    \includegraphics[width=1.0\linewidth]{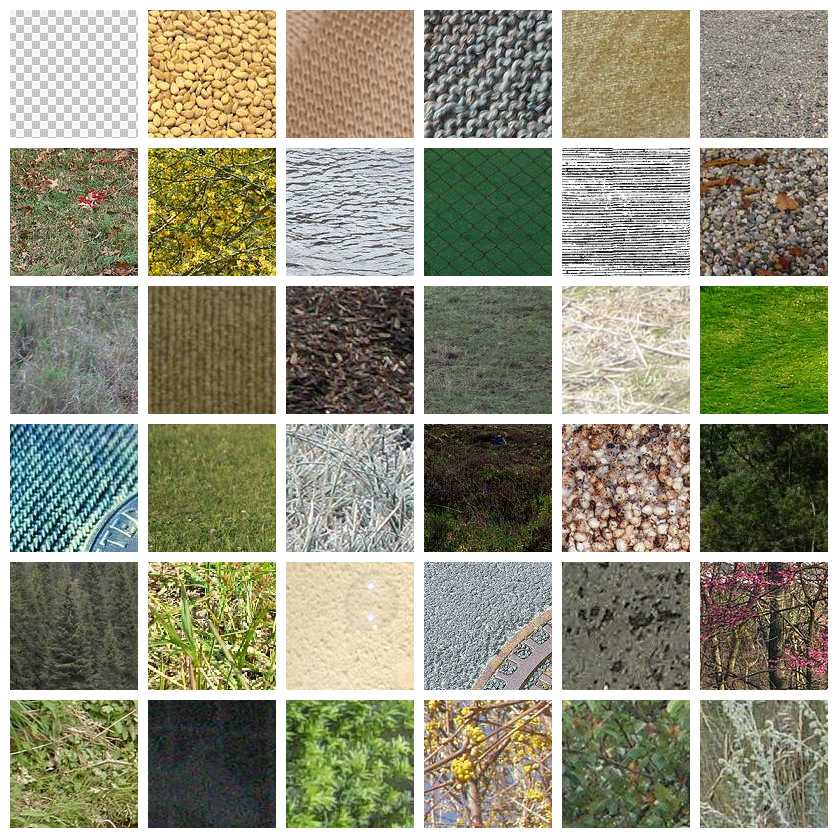}
    \caption{Example images from our training dataset, ordered from most to least homogeneous according to the score, in equal steps of ranking.}
    \label{fig:dataset}
\end{figure}

\begin{figure}[h]
    \centering
    \begin{minipage}[t]{0.03\linewidth}
    \vspace{30pt}
    (a) \\[52pt]
    (b) \\[52pt]
    (c) \\[52pt]
    (d) \\[52pt]
    (e) \\[52pt]
    (f)
    \end{minipage}
    \begin{minipage}[t]{0.96\linewidth}
    \vspace{0pt}
    \includegraphics[width=1.0\linewidth]{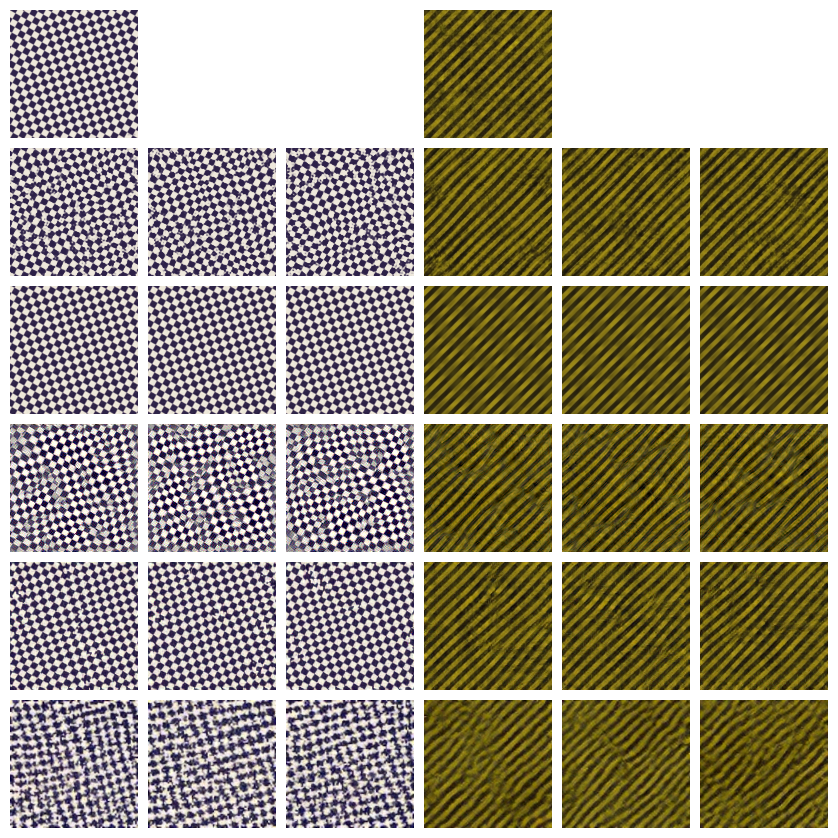}
    \end{minipage}
    
    \caption{Resampling with three samples per texture image per method.
    \textbf{(a)} Original texture images.
    \textbf{(b)} Portilla-Simoncelli model;
    \textbf{(c)} $\mu$NCA model;
    \textbf{(d)} Gatys model;
    \textbf{(e)} statistics matching for our model;
    \textbf{(f)} diffusion sampling for our model.}
    \label{fig:resample-repeat-1}
\end{figure}

\begin{figure}[h]
    \centering
    \begin{minipage}[t]{0.03\linewidth}
    \vspace{30pt}
    (a) \\[52pt]
    (b) \\[52pt]
    (c) \\[52pt]
    (d) \\[52pt]
    (e) \\[52pt]
    (f)
    \end{minipage}
    \begin{minipage}[t]{0.96\linewidth}
    \vspace{0pt}
    \includegraphics[width=1.0\linewidth]{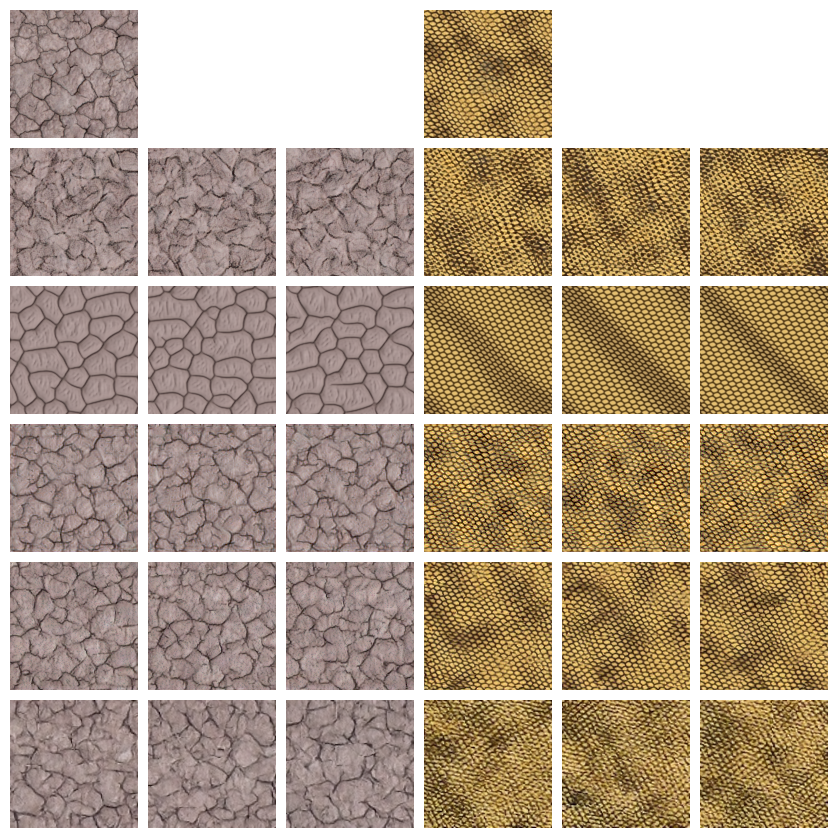}
    \end{minipage}
    
    \caption{Resampling with three samples per texture image per method.
    \textbf{(a)} Original texture images.
    \textbf{(b)} Portilla-Simoncelli model;
    \textbf{(c)} $\mu$NCA model;
    \textbf{(d)} Gatys model;
    \textbf{(e)} statistics matching for our model;
    \textbf{(f)} diffusion sampling for our model.}
    \label{fig:resample-repeat-2}
\end{figure}

\begin{figure}
    \centering
    \begin{minipage}[t]{0.03\linewidth}
    \vspace{30pt}
    (a) \\[52pt]
    (b) \\[52pt]
    (c) \\[52pt]
    (d) \\[52pt]
    (e) \\[52pt]
    (f)
    \end{minipage}
    \begin{minipage}[t]{0.96\linewidth}
    \vspace{0pt}
    \includegraphics[width=1.0\linewidth]{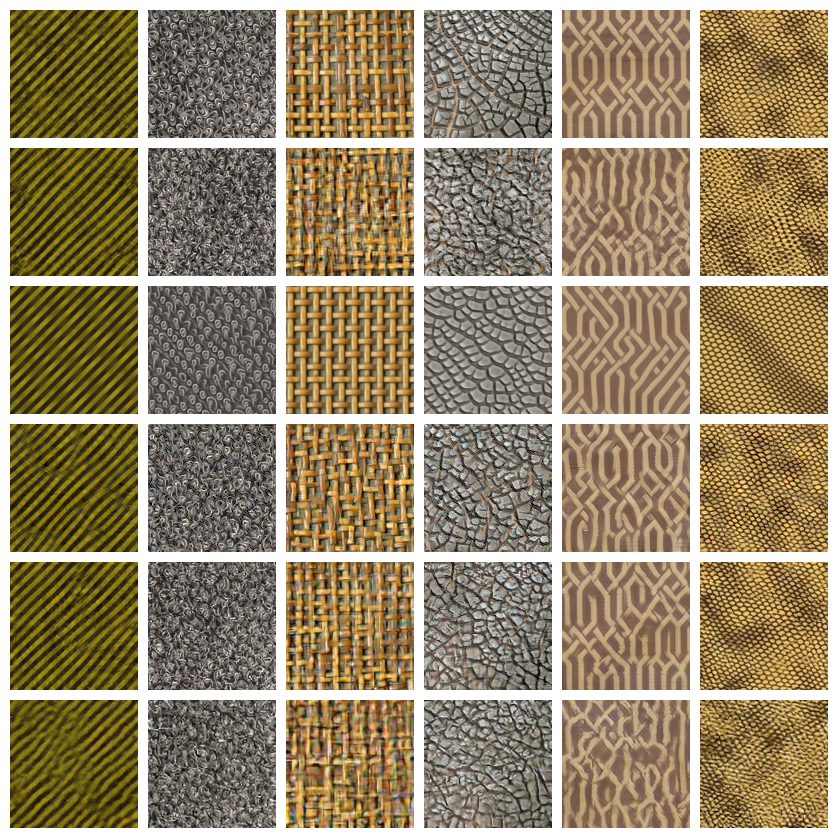}
    \end{minipage}
    
    \caption{Resampling examples.
    \textbf{(a)} Original texture images (from DTD dataset, sorted in decreasing homogeneity).  Remaining rows show samples generated by 
    \textbf{(b)} Portilla-Simoncelli model;
    \textbf{(c)} $\mu$NCA model;
    \textbf{(d)} Gatys model;
    \textbf{(e)} statistics matching for our model;
    \textbf{(f)} diffusion sampling for our model. }
    \label{fig:resample2}
\end{figure}

\begin{figure}
    \centering
    \begin{minipage}[t]{0.03\linewidth}
    \vspace{30pt}
    (a) \\[52pt]
    (b) \\[52pt]
    (c) \\[52pt]
    (d) \\[52pt]
    (e) \\[52pt]
    (f)
    \end{minipage}
    \begin{minipage}[t]{0.96\linewidth}
    \vspace{0pt}
    \includegraphics[width=1.0\linewidth]{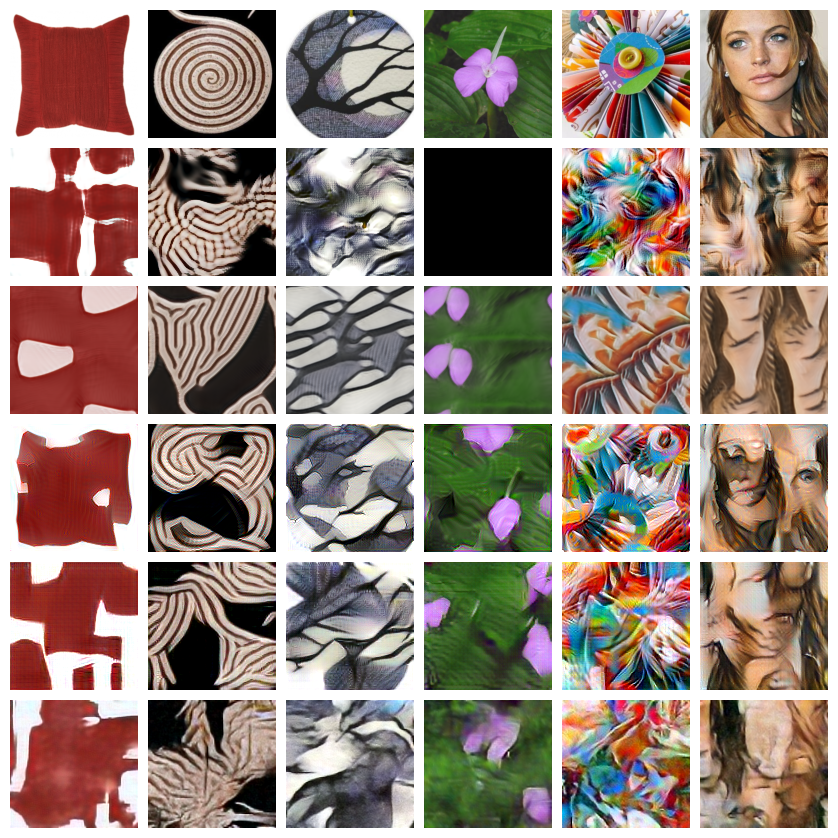}
    \end{minipage}
    
    \caption{Resampling examples for inhomogeneous images.
    \textbf{(a)} Original texture images (from DTD dataset).  Remaining rows show samples generated by 
    \textbf{(b)} Portilla-Simoncelli model (a synthesis failure occurred in the fourth panel);
    \textbf{(c)} $\mu$NCA model;
    \textbf{(d)} Gatys model;
    \textbf{(e)} statistics matching for our model;
    \textbf{(f)} diffusion sampling for our model. }
    \label{fig:resample-inhomogeneous}
\end{figure}

\begin{figure}[h]
\centering
    \begin{minipage}[t]{0.03\linewidth}
    \vspace{30pt}
    (a) \\[52pt]
    (b) \\[52pt]
    (c) \\[52pt]
    (d)
    \end{minipage}
    \begin{minipage}[t]{0.96\linewidth}
    \vspace{0pt}
    \includegraphics[width=1.0\linewidth]{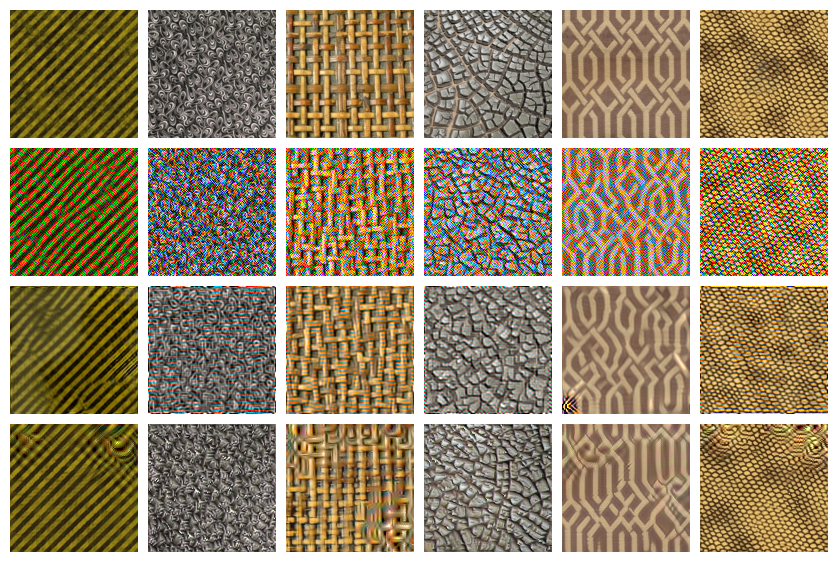}
    \end{minipage}
    
    \caption{Competitive adversarial comparison between the Gatys model and our model.
    \textbf{(a)} Original texture images (same as Figure \ref{fig:resample2}).
    \textbf{(b)} Images that are considered the same as (a) according to the Gatys model and maximally different according to our model (zoom in to see the high frequency details).
    \textbf{(c)} Same as (b) but adding a high-pass statistic to the Gatys model.
    \textbf{(d)} Images that are considered the same as (a) according to our model and maximally different according to the Gatys model.}
    \label{fig:competition2}
\end{figure}

\begin{figure}[h]
    \centering
    \begin{minipage}[t]{0.03\linewidth}
    \vspace{30pt}
    (a) \\[52pt]
    (b) \\[52pt]
    (c) \\[52pt]
    (d) \\[52pt]
    (e)
    \end{minipage}
    \begin{minipage}[t]{0.96\linewidth}
    \vspace{0pt}
    \includegraphics[width=1.0\linewidth]{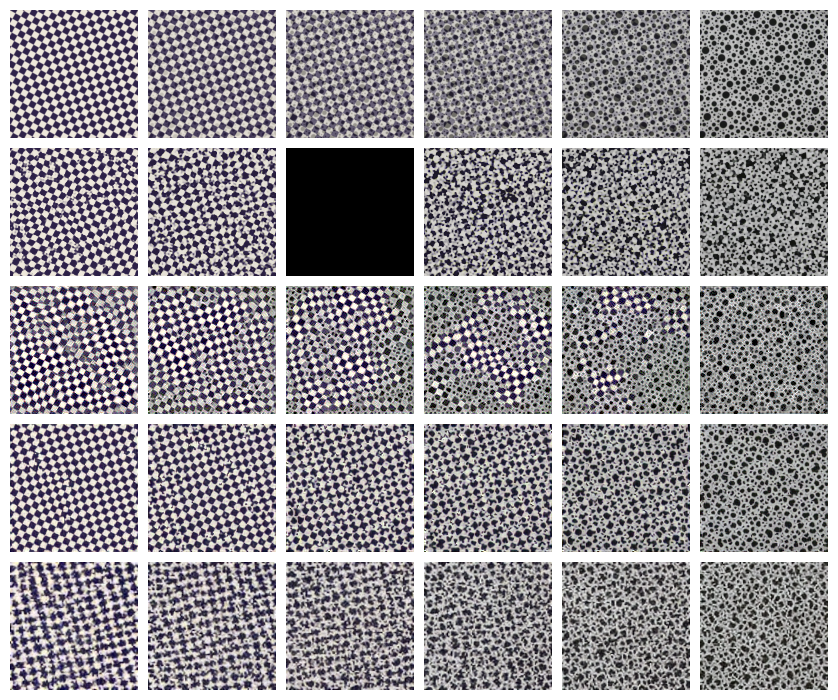}
    \end{minipage}
    
    \caption{Interpolation between 1st and 2nd images in Figure \ref{fig:resample}. \textbf{(a)} Fading in pixel values. \textbf{(b)} Portilla-Simoncelli model (a synthesis failure occurred in the third panel). \textbf{(c)} Gatys model. \textbf{(d)} Statistics matching for our model. \textbf{(e)} Diffusion sampling for our model.}
    \label{fig:interpolate2}
\end{figure}

\begin{figure}[h]
    \centering
    \begin{minipage}[t]{0.03\linewidth}
    \vspace{30pt}
    (a) \\[52pt]
    (b) \\[52pt]
    (c) \\[52pt]
    (d) \\[52pt]
    (e)
    \end{minipage}
    \begin{minipage}[t]{0.96\linewidth}
    \vspace{0pt}
    \includegraphics[width=1.0\linewidth]{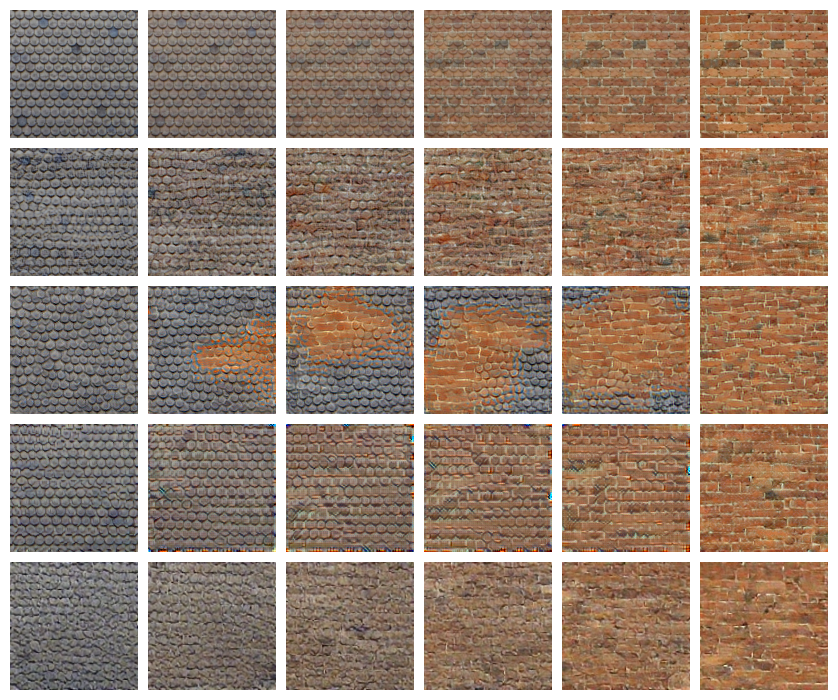}
    \end{minipage}
    
    \caption{Interpolation between 3rd and 4th images in Figure \ref{fig:resample}. \textbf{(a)} Fading in pixel values. \textbf{(b)} Portilla-Simoncelli model. \textbf{(c)} Gatys model. \textbf{(d)} Statistics matching for our model. \textbf{(e)} Diffusion sampling for our model.}
    \label{fig:interpolate3}
\end{figure}

Figure \ref{fig:dataset} provides example images from our training dataset, sorted according to their homogeneity score.
Figures \ref{fig:resample-repeat-1} and \ref{fig:resample-repeat-2} show multiple samples per texture image per method. Figure \ref{fig:resample2} shows resampling for more texture images (in addition to Figure \ref{fig:resample}). Figure \ref{fig:resample-inhomogeneous} shows resampling for inhomogeneous images. Figure \ref{fig:competition2} shows competitive adversarial comparison for more texture images (in addition to Figure \ref{fig:competition}). Figures \ref{fig:interpolate2} and \ref{fig:interpolate3} provide additional examples of interpolation (in addition to Figure \ref{fig:interpolate}).

\clearpage

\end{document}